\documentclass[letterpaper]{article} % DO NOT CHANGE THIS
\usepackage{aaai2027}  % DO NOT CHANGE THIS
\usepackage[hyphens]{url}  % DO NOT CHANGE THIS
\usepackage{graphicx} % DO NOT CHANGE THIS
\usepackage{natbib}  % DO NOT CHANGE THIS AND DO NOT ADD ANY OPTIONS TO IT
\usepackage{caption} % DO NOT CHANGE THIS AND DO NOT ADD ANY OPTIONS TO IT
\usepackage{algorithm}
\usepackage{algorithmic}

\usepackage{newfloat}
\usepackage{listings}
\DeclareCaptionStyle{ruled}{labelfont=normalfont,labelsep=colon,strut=off} % DO NOT CHANGE THIS
\floatstyle{ruled}
\newfloat{listing}{tb}{lst}{}
\floatname{listing}{Listing}

\usepackage{booktabs}

\usepackage{booktabs}
\usepackage{multirow}
\usepackage{graphicx}
\usepackage{arydshln}
\usepackage{amssymb}
\usepackage{amsmath}
\usepackage[table]{xcolor}

\newcommand\ie{{\it i.e.}}
\newcommand\eg{{\it e.g.}}
\title{ProPRL: Property-Aware Prerequisite Relation Learning in Educational Knowledge Graphs}
\author{
    Xinghe Cheng\textsuperscript{\rm 1},
    Jiapu Wang\textsuperscript{\rm 2},
    Chaobo He\textsuperscript{\rm 3},
    Ruihai Dong\textsuperscript{\rm 4},
    Quanlong Guan\textsuperscript{\rm 1}
}
\affiliations{
    \textsuperscript{\rm 1}Jinan University\\
    \textsuperscript{\rm 2}Nanjing University of Science and Technology\\
    \textsuperscript{\rm 3}South China Normal University\\
    \textsuperscript{\rm 3}University College Dublin\\
    chengxh@stu2023.jnu.edu.cn, jiapuwang9@gmail.com, hechaobo@m.scnu.edu.cn\\
    gql@jnu.edu.cn, ruihai.dong@ucd.ie
}

\begin{document}

\maketitle

\begin{abstract}
Prerequisite relation learning is central to adaptive instruction, yet existing methods often formulate it as conventional link prediction, limiting their ability to adaptively integrate complementary educational evidence for individual candidate pairs and to discourage contradictory reverse predictions. We propose \textbf{ProPRL}, a \textbf{Pro}perty-aware \textbf{P}rerequisite \textbf{R}elation \textbf{L}earning framework. ProPRL first learns complementary concept representations from a concept-resource hypergraph and a directed learning-behavior graph, where direction-preserving personalized propagation aggregates multi-hop behavioral evidence. It then employs a \textit{Pair-conditioned Gate} to adaptively weight and fuse the two views for each candidate ordered concept pair. Finally, an \textit{Irreversibility Constraint} introduces an anti-symmetry regularizer that penalizes simultaneously high confidence in both directions of the same concept pair. Experiments on multiple real-world educational datasets show that ProPRL achieves state-of-the-art performance on prerequisite relation learning.
\end{abstract}

\section{Introduction}
\looseness=-1
As a core pillar of smart education, personalized learning fundamentally relies on accurate domain knowledge structures to facilitate adaptive instructional decision-making~\cite{CheZF2025,AbdWN2023}.
Within these structures, prerequisite relations serve as foundational components that map the hierarchical dependencies between knowledge concepts~\cite{ZhaWH2025}.
However, modeling these relations in real-world educational settings is often hindered by the scarcity of reliable expert annotations and the inherent noise in automated extraction methods~\cite{CheZW2026}.
Prerequisite relation learning aims to mitigate these challenges by inferring implicit dependencies from diverse educational data, thereby establishing a more robust structural basis for effective, adaptive instruction.

\looseness=-1
The methodological paradigm of prerequisite relation learning has undergone a definitive transition from feature-centric heuristics to expressive graph representations~\cite{LiaWH2015,MazPB2023}.
Historically, early studies mainly relied on manually designed features, textual links, or course dependency structures, applying these metrics widely to Wikipedia, university curricula, and MOOC resources~\cite{HuXW2021}.
To capture more intricate dependencies, current models shift towards complex topologies, utilizing heterogeneous graph neural networks to map relations across concepts and resources.
These frameworks increasingly integrate multi-view structures and learning paths to characterize dependencies from broader data dimensions~\cite{SunLZ2022,ZhaWX2025}.

\begin{figure}[t]
  \centering
  \includegraphics[width=\linewidth]{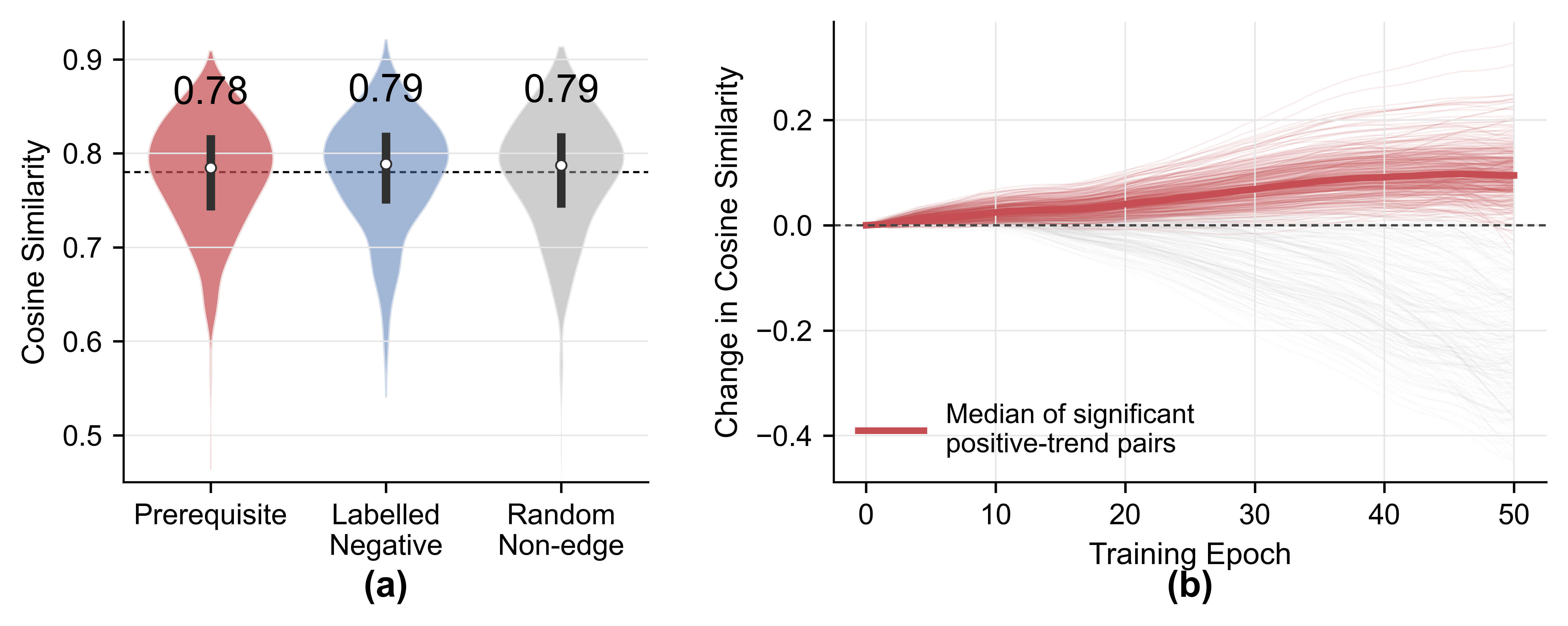}
  \caption{Representation homogenization of knowledge concepts. Left: Concepts exhibit high similarity in the pretrained semantic space. Right: During DGCPL~\cite{ZhaWH2025} training, 62.7\% of prerequisite pairs become increasingly similar.}
  \label{fig:motivation}
\end{figure}

Despite these advances, most existing methods reduce prerequisite relation prediction to conventional link prediction and infer relations primarily from embedding proximity.
As shown in Figure \ref{fig:motivation}(a), prerequisite pairs, labeled negative pairs, and random non-edges exhibit similarly high initial semantic similarities, making them difficult to distinguish. 
Figure \ref{fig:motivation}(b) further shows that GNN training significantly increases the similarity of many concept pairs, progressively homogenizing their representations.
These observations motivate the explicit modeling of three important characteristics: directional asymmetry, multi-hop behavioral evidence, and pair-specific relevance.

\looseness=-1
First, prerequisite relations are intrinsically asymmetric and directionally irreversible~\cite{AlzKP2018,YanGM2024}.
When concept $c_i$ is a prerequisite of $c_j$, assigning high confidence to the reverse relation $c_j\rightarrow c_i$ would introduce a directional contradiction.
Yet, conventional link prediction models evaluate these directed pairs independently, failing to explicitly penalize such inverse combinations.
This fundamental misalignment frequently produces contradictory bidirectional predictions, thereby degrading the directional consistency of the graph and inherently limiting the task of accurately resolving asymmetric conceptual dependencies.

\looseness=-1
Second, prerequisite relations often manifest as latent multi-hop dependencies rather than explicit one-step transitions in learner behaviors~\cite{TalC2012,XiaBZ2022}.
Even when concepts $c_i$ and $c_j$ are not studied consecutively, intermediate interactions (\eg, $c_i \rightarrow c_k \rightarrow c_j$) provide vital indirect prerequisite evidence.
Yet, conventional behavior modeling frameworks are largely tethered to local transition heuristics.
This shortsightedness prevents them from effectively propagating structural signals across the behavioral graph, thereby losing the long-range dependencies required to accurately reconstruct the underlying knowledge topology.

\looseness=-1
Third, prerequisite relations are inherently defined over specific concept pairs rather than isolated concepts.
Existing graph-based methods typically model concepts independently, learning global node representations before applying a generic scoring function~\cite{ZhaWH2025}.
Under this node-centric paradigm, the representation of $c_i$ remains fixed across its pairings with $c_j$ and $c_k$.
It therefore cannot adapt to the distinct relational context provided by each candidate concept.
Prerequisite relation learning thus requires pair-conditioned relation reasoning that constructs dependency evidence tailored to each specific concept pair.

\looseness=-1
To address these limitations, we propose \textbf{ProPRL}, a property-aware framework for prerequisite relation learning.
ProPRL jointly exploits complementary structural and behavioral evidence, adapts the contribution of these evidence sources to each candidate relation, and discourages mutually contradictory predictions in opposite directions.
Specifically, it integrates concept representations from concept-resource and learning-behavior views, employs a pair-conditioned gate to assign relation-specific importance to the two views, and incorporates an anti-symmetry regularizer into the training objective.
Extensive experiments on multiple educational datasets demonstrate that ProPRL outperforms existing baselines.

\looseness=-1
The main contributions of this work are as follows:
\begin{itemize}
    \item We introduce \textbf{ProPRL}, a property-aware framework for prerequisite relation learning that explicitly captures complementary multi-view evidence, pair-specific relevance, and directional irreversibility.
    \item To the best of our knowledge, we are the first to introduce direction-preserving personalized propagation over directed learning-behavior graphs for prerequisite relation learning.
    \item We develop a Pair-conditioned Gate that constructs view-specific representations for each candidate ordered concept pair and adaptively weights and fuses the two evidence views according to their pair-specific relevance. We further introduce an irreversibility-aware anti-symmetry regularizer that discourages simultaneously high confidence in both directions of the same concept pair.
\end{itemize}

\section{Related Work}
\paragraph{\textbf{Feature-Based Concept Prerequisite Relation Learning.}}
Concept prerequisite relation learning aims to identify directed dependencies between knowledge concepts, which is fundamental to curriculum planning, learning path recommendation, and intelligent tutoring~\cite{LiaWH2015,RoyML2019}.
Early studies mainly formulated this task as a feature-based prediction problem.
For example, RefD measures prerequisite relations by exploiting asymmetric reference patterns between Wikipedia concepts~\cite{LiaWH2015}, while course dependency information has been used to recover concept-level prerequisite relations from university curricula~\cite{LiaYW2017}.
With the growth of online educational resources, subsequent studies learned prerequisite relations from MOOCs, lecture materials, textbooks, and online resources~\cite{LiFT2019,RoyML2019}.
These methods improve scalability by learning concept representations or concept-pair features from educational content.
However, they mainly rely on static content, course structures, or resource dependencies, and thus provide limited modeling of prerequisite evidence emerging from learners' actual learning processes~\cite{ZhaWH2025}.

\paragraph{\textbf{Graph-based Concept Prerequisite Relation Learning.}}
Recent studies increasingly model prerequisite relation learning as a graph-based relation prediction problem~\cite{ZhaLZ2022,MazPB2023}.
R-VGAE introduces a relational variational graph autoencoder to learn prerequisite chains without labeled concept pairs~\cite{LiFH2020}.
Heterogeneous graph models further incorporate concept-resource or learning-object dependencies to improve concept representation learning~\cite{MazPB2023}.
MHA-VGAE models interactions between resource graphs and concept graphs with multi-head attention and variational graph autoencoders~\cite{ZhaLZ2022}.
More recent work incorporates learning paths, global knowledge relations, or multi-view graph structures to capture richer dependency signals~\cite{SunHX2024,ZhaWH2025}.
LCPRE~\cite{SunHX2024} uses learning-path supervision to extract prerequisite relations from educational data, GKROM~\cite{ZhaWX2025} optimizes multiple types of knowledge relations for prerequisite relation learning, and DGCPL constructs both a concept-resource hypergraph and a learner behavior graph to integrate knowledge and behavior perspectives~\cite{ZhaWH2025}.
Nevertheless, most graph-based methods still learn concept-level representations and predict candidate edges independently, leaving ordered concept-pair reasoning and directional irreversibility insufficiently explored.

\begin{figure*}[t]
  \centering
  \includegraphics[width=\linewidth]{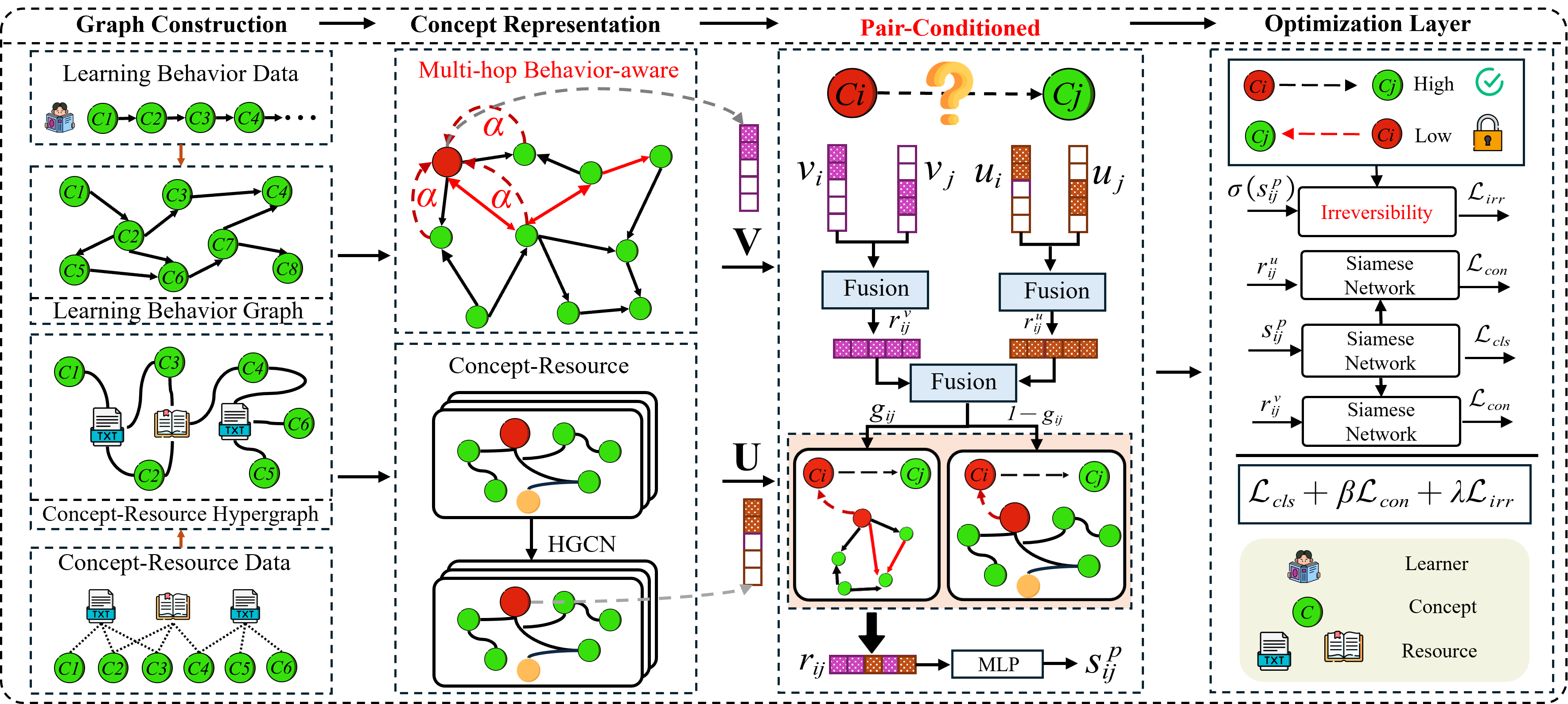}
  \caption{Overview of the ProPRL framework. ProPRL comprises four stages. (a) Graph construction organizes concept–resource associations and learner interactions into two complementary graph views. (b) Concept representation captures high-order structural dependencies and multi-hop behavioral information. (c) Pair-conditioned fusion dynamically integrates the two views to generate a pair-specific embedding for each candidate concept pair. (d) Constraint-aware optimization learns prerequisite relations while suppressing contradictory reverse predictions.}
  \label{fig:framework}
\end{figure*}

\section{Problem Definition}
\looseness=-1
Let $\mathcal{D}$ denote an educational dataset comprising a concept set $\mathcal{C}=\{c_1,c_2,\ldots,c_n\}$, a resource set $\mathcal{R}=\{r_1,r_2,\ldots,r_m\}$, a learner set $\mathcal{U}$, learner interaction sequences $\mathcal{S}$, and observed prerequisite labels $\mathcal{Y}$.
Specifically, $\mathcal{S}=\{\mathbf{s}_u \mid u\in\mathcal{U}\}$, where $\mathbf{s}_u=(c_{u,1},c_{u,2},\ldots,c_{u,T_u})$ denotes the interaction sequence of learner $u$, with $c_{u,t}\in\mathcal{C}$ representing the concept involved in the $t$-th interaction.
Each concept $c_i \in \mathcal{C}$ is associated with an initial feature vector $\mathbf{x}_i\in\mathbb{R}^{d}$, and all concept features are organized as a matrix $\mathbf{X}\in\mathbb{R}^{n\times d}$.

For each labeled ordered concept pair $(c_i,c_j)$, where $i\neq j$, let $y_{ij}\in\{0,1\}$ denote its observed label. Specifically, $y_{ij}=1$ if $c_i$ is a prerequisite of $c_j$, and $y_{ij}=0$ otherwise. The collection of all observed labels is denoted by $\mathcal{Y}$.
Given $\mathcal{D}$, our goal is to learn a property-aware prerequisite predictor $\mathcal{F}$ such that:
\begin{equation}
    \mathcal{F}(c_i,c_j \mid \mathcal{D}) = p_{ij},
\end{equation}
where $p_{ij}\in[0,1]$ denotes the predicted probability that $c_i$ is a prerequisite of $c_j$.

\section{Method}
\looseness=-1
We propose \textbf{ProPRL}, a \textbf{Pro}perty-aware framework for \textbf{P}rerequisite \textbf{R}elation L\textbf{}earning.
ProPRL is built around three principal components: \textit{Multi-view Concept Representation}, \textit{Pair-conditioned Gate}, and \textit{Irreversibility Constraint}.
\textit{Multi-view Concept Representation} learns complementary concept embeddings from a directed learning-behavior graph and a concept-resource hypergraph.
The behavior view employs direction-preserving personalized multi-hop propagation to aggregate multi-hop behavioral evidence, whereas the concept-resource view captures resource-mediated high-order concept associations.
Building on these representations, \textit{Pair-conditioned Gate} constructs view-specific representations for each candidate ordered concept pair and learns a pair-specific gate to adaptively weight and fuse evidence from the two views.
This pair-conditioned fusion allows the contribution of each view to vary across candidate pairs, rather than relying on a fixed node-level fusion.
Finally, \textit{Irreversibility Constraint} introduces an anti-symmetry regularizer that penalizes simultaneously high predictions for $(c_i,c_j)$ and $(c_j,c_i)$, thereby encouraging directional consistency in prerequisite predictions.

\paragraph{\textbf{Multi-view Concept Representation.}}
%\noindent \textbf{Behavior-aware Multi-hop Propagation}
%\looseness=-1
We first construct a directed learning behavior graph $\mathcal{G}_B$ from learner interaction sequences $\mathcal{S}$ to capture the transitional dynamics between concepts. 
If a sequential transition from concept $c_i$ to $c_j$ is observed, a directed edge $c_i \rightarrow c_j$ is established. 
Let $\mathbf{B}^{\mathrm{out}}\in\mathbb{R}^{n\times n}$ denote the weighted outgoing adjacency matrix, where $\mathbf{B}^{\mathrm{out}}_{ij}$ quantifies the transition strength from $c_i$ to $c_j$. 
Conversely, the incoming adjacency matrix is defined as $\mathbf{B}^{\mathrm{in}}=(\mathbf{B}^{\mathrm{out}})^{\top}$, explicitly capturing the reverse behavioral flow.

To mitigate numerical instability during graph propagation, we incorporate self-loops prior to normalization. 
For each direction $d\in\{\mathrm{out},\mathrm{in}\}$, we define $\widetilde{\mathbf{B}}^{d}=\mathbf{B}^{d}+\mathbf{I}$ and its corresponding diagonal degree matrix $\mathbf{D}^{d}$, where
\begin{equation}
    \mathbf{D}^{d}_{ii} = \sum_{j=1}^{n}\widetilde{\mathbf{B}}^{d}_{ij}.
\end{equation}
The normalized behavior adjacency matrices are computed as:
\begin{equation}
    \mathbf{A}^{d} = (\mathbf{D}^{d})^{-1}\widetilde{\mathbf{B}}^{d}.
\end{equation}
Thus, $\mathbf{A}^{\mathrm{out}}$ and $\mathbf{A}^{\mathrm{in}}$ serve as the fundamental topologies for aggregating forward and backward transitional evidence, respectively.

To encode localized structural contexts, we deploy direction-specific graph convolutional networks (GCNs)~\cite{KipWM2016} equipped with shortcut transformations. 
The $L$-layer message passing is formulated as:
\begin{equation}
    \mathbf{H}_{d}^{(\ell+1)} = \phi\left(\mathbf{A}^{d}\mathbf{H}_{d}^{(\ell)}\mathbf{W}_{g,d}^{(\ell)}\right) + \mathbf{H}_{d}^{(\ell)} \mathbf{W}_{0,d}^{(\ell)},
\end{equation}
\begin{equation}
    \mathbf{H}_{d}^{(0)}=\mathbf{X},
\end{equation}
where $\mathbf{W}_{g,d}^{(\ell)}$ and $\mathbf{W}_{0,d}^{(\ell)}$ are trainable weight matrices, $\phi(\cdot)$ is a nonlinear activation function. 
After $L$ layers, the outgoing and incoming behavioral representations are extracted as $\mathbf{Q}^{\mathrm{out}}=\mathbf{H}_{\mathrm{out}}^{(L)}$ and $\mathbf{Q}^{\mathrm{in}}=\mathbf{H}_{\mathrm{in}}^{(L)}$.

As prerequisite relationships frequently manifest as latent multi-hop dependencies rather than strictly adjacent transitions, motivated by the success of APPNP~\cite{GasBG2018} in global topological exploration, we devise a specialized propagation scheme to distill long-range structural signals.
The iterative propagation is defined as:
\begin{equation}
    \mathbf{Z}^{d,(t+1)} = (1-\alpha)\mathbf{A}^{d}\mathbf{Z}^{d,(t)} + \alpha\mathbf{Q}^{d},
    \quad
    \mathbf{Z}^{d,(0)}=\mathbf{Q}^{d},
\end{equation}
where $t=0,1,\ldots,k-1$ indexes the propagation step, and $\alpha\in(0,1]$ acts as a teleport probability that gracefully balances higher-order neighborhood exploration with the preservation of localized concept semantics. 
After $k$ steps, the fully propagated representations are denoted as $\mathbf{Z}^{\mathrm{out}}=\mathbf{Z}^{\mathrm{out},(k)}$ and $\mathbf{Z}^{\mathrm{in}}=\mathbf{Z}^{\mathrm{in},(k)}$.

Finally, to holistically represent the asymmetric interaction profile of each concept, we fuse the directional representations:
\begin{equation}
    \mathbf{V} = \operatorname{LN} \left(\phi\left(\mathbf{Z}^{\mathrm{out}}\mathbf{W}_{\mathrm{out}} + \mathbf{Z}^{\mathrm{in}}\mathbf{W}_{\mathrm{in}} \right) \right),
\end{equation}
where $\mathbf{W}_{\mathrm{out}}$ and $\mathbf{W}_{\mathrm{in}}$ are trainable projection matrices, and $\operatorname{LN}(\cdot)$ denotes layer normalization. 
This yields the final behavior-aware representation matrix $\mathbf{V}$, with $\mathbf{v}_i$ denoting the behavior-aware representation for concept $c_i$.

%\subsubsection{Resource-aware Concept Context Encoding}
\looseness=-1
We construct a concept-resource hypergraph and employ a Hypergraph Convolutional Network (HGCN)~\cite{FenYZ2019} to capture high-order associations between concepts and learning resources.
Let $\mathbf{M}\in\mathbb{R}^{n\times m}$ denote the concept-resource incidence matrix, where $\mathbf{M}_{ij}$ indicates whether concept $c_i$ is associated with resource $r_j$.
Each resource is treated as a hyperedge connecting its related concepts.

Based on $\mathbf{M}$, we define the concept degree matrix $\mathbf{D}_{c}\in\mathbb{R}^{n\times n}$ and the resource degree matrix $\mathbf{D}_{r}\in\mathbb{R}^{m\times m}$.
Their diagonal entries are computed as:
\begin{equation}
    (\mathbf{D}_{c})_{ii} = \sum_{j=1}^{m} \mathbf{M}_{ij}\omega_j,
    \quad
    (\mathbf{D}_{r})_{jj} = \sum_{i=1}^{n} \mathbf{M}_{ij},
\end{equation}
where $\omega_j$ is the weight of resource hyperedge $r_j$.
In our implementation, all resource hyperedges are assigned equal weights, \ie, $\omega_j=1$.
Let $\mathbf{\Omega}=\operatorname{diag}(\omega_1,\omega_2,\ldots,\omega_m)$.
The normalized hypergraph propagation matrix is then defined as:
\begin{equation}
    \mathbf{P} = \mathbf{D}_{c}^{-\frac{1}{2}} \mathbf{M} \mathbf{\Omega} \mathbf{D}_{r}^{-1} \mathbf{M}^{\top} \mathbf{D}_{c}^{-\frac{1}{2}}.
\end{equation}
This normalization first propagates information from concepts to resource hyperedges and then back to concepts, while accounting for both concept degrees and resource hyperedge sizes.

Starting from $\mathbf{U}^{(0)}=\mathbf{X}$, the resource-aware encoder updates concept representations by:
\begin{equation}
    \mathbf{U}^{(\ell+1)} = \phi\left( \mathbf{P}\mathbf{U}^{(\ell)}\mathbf{W}_{h}^{(\ell)} \right) + \mathbf{U}^{(\ell)}\mathbf{W}_{0}^{(\ell)},
\end{equation} 
where $\mathbf{U}^{(\ell)}$ is the concept representation matrix at layer $\ell$, $\mathbf{W}_{h}^{(\ell)}$ is the trainable weight matrix of the hypergraph aggregation branch and $\mathbf{W}_{0}^{(\ell)}$ is the trainable weight matrix of the shortcut transformation branch.

\paragraph{\textbf{Pair-conditioned Gate.}}
Prerequisite completion is defined over ordered concept pairs rather than individual concepts. Although the resource-aware representation $\mathbf{u}_i$ and the behavior-aware representation $\mathbf{v}_i$ encode concept-level evidence, they do not by themselves specify the role of a concept in a candidate relation. The same concept may serve as a prerequisite in one pair but as a target concept in another pair. Therefore, ProPRL constructs pair-conditioned representations to model the candidate relation from $c_i$ to $c_j$.

\looseness=-1
For the resource-aware view, we compose the representations of the source concept $c_i$ and the target concept $c_j$ as:
\begin{equation}
    \mathbf{r}_{ij}^{u} = [\mathbf{u}_i;\mathbf{u}_j;\mathbf{u}_i-\mathbf{u}_j; \mathbf{u}_i\odot\mathbf{u}_j],
\end{equation}
where $[\cdot;\cdot]$ denotes vector concatenation and $\odot$ denotes the Hadamard product. The first two terms preserve the role-specific information of the source and target concepts.
The signed difference $\mathbf{u}_i-\mathbf{u}_j$ introduces order-sensitive evidence, since reversing the pair changes this term. The product term $\mathbf{u}_i\odot\mathbf{u}_j$ captures dimension-wise compatibility between the two concepts.

\looseness=-1
Following a parallel logic, the behavior-aware pair representation is defined as:
\begin{equation}
    \mathbf{r}_{ij}^{v} = [\mathbf{v}_i;\mathbf{v}_j;\mathbf{v}_i-\mathbf{v}_j; \mathbf{v}_i\odot\mathbf{v}_j].
\end{equation}
This representation describes the candidate pair from the perspective of learner behavior, including both directional behavioral discrepancy and pairwise behavioral compatibility.

\looseness=-1
The resource-aware and behavior-aware views may contribute differently to different candidate pairs. For example, some prerequisite relations can be supported mainly by shared learning resources, whereas others may be more clearly reflected in sequential learning behavior. A fixed view-level fusion strategy would ignore such candidate-specific differences. To address this issue, ProPRL learns a pair-conditioned gate:
\begin{equation}
    \mathbf{g}_{ij} = \sigma\left(\operatorname{MLP}_{g}\left([\mathbf{r}_{ij}^{u};\mathbf{r}_{ij}^{v};|\mathbf{r}_{ij}^{u}-\mathbf{r}_{ij}^{v}|;\mathbf{r}_{ij}^{u}\odot\mathbf{r}_{ij}^{v}]\right)\right),
\end{equation}
where \(\sigma(\cdot)\) denotes the sigmoid function and $\mathbf{g}_{ij}$ has the same dimension as $\mathbf{r}_{ij}^{u}$ and $\mathbf{r}_{ij}^{v}$.
The absolute difference term measures the discrepancy between the two views, while the Hadamard product captures their dimension-wise agreement. The gate therefore determines, for each candidate pair and each feature dimension, how much information should be taken from the resource-aware view or the behavior-aware view.

\looseness=-1
The final pair representation is obtained by gated fusion:
\begin{equation}
    \mathbf{r}_{ij} = \mathbf{g}_{ij}\odot\mathbf{r}_{ij}^{u} + (1-\mathbf{g}_{ij})\odot\mathbf{r}_{ij}^{v}.
\end{equation}
Based on $\mathbf{r}_{ij}$, the prerequisite score and probability are computed as:
\begin{equation}
    s_{ij}^{p} = \operatorname{MLP}_{p}(\mathbf{r}_{ij}),
    \quad
    p_{ij}^{p} = \sigma(s_{ij}^{p}),
\end{equation}
where $p_{ij}^{p}$ denotes the fused prerequisite probability that $c_i$ is a prerequisite of $c_j$.

\paragraph{\textbf{Irreversibility Constraint.}}
\looseness=-1
Prerequisite relations are directional and generally irreversible.
If concept $c_i$ is a prerequisite of concept $c_j$, the reverse relation from $c_j$ to $c_i$ should not be assigned a high probability at the same time.
However, when ordered pairs are optimized independently, a model may still produce contradictory bidirectional predictions.
To reduce such structural inconsistency, ProPRL introduces an order-sensitive irreversibility constraint.

For each positive training pair $(c_i,c_j)$, ProPRL additionally evaluates its reverse pair $(c_j,c_i)$ using the same pair scoring function.
Let $p_{ij}^{p}$ and $p_{ji}^{p}$ denote the fused prerequisite probabilities of the forward and reverse directions, respectively.
The irreversibility loss is defined as:
\begin{equation}
    \mathcal{L}_{\mathrm{irr}} = \frac{1}{|\mathcal{T}^{+}|} \sum_{(i,j)\in\mathcal{T}^{+}} \max\left(0, p_{ij}^{p}+p_{ji}^{p}-\mu\right),
\end{equation}
where $\mathcal{T}^{+}=\{(i,j)\in\mathcal{T}\mid y_{ij}=1\}$ denotes the set of positive training pairs, and $\mu$ is a co-activation margin that controls the allowed upper bound of the two opposite-direction probabilities.

This constraint penalizes cases in which both directions receive high probabilities. Therefore, it encourages the model to preserve the forward prerequisite evidence while suppressing the reverse prediction, leading to more directionally consistent prerequisite completion.

\paragraph{\textbf{Overall Optimization.}}
\looseness=-1
ProPRL optimizes three prediction branches jointly: a resource-aware branch, a behavior-aware branch, and a pair-aware fused branch. The resource-aware and behavior-aware branches are supervised not only to provide auxiliary predictions, but also to preserve view-specific prerequisite evidence.
For each view $a\in\{u,v\}$, we use a Siamese relation classifier to score an ordered pair.
Specifically, the two concepts in a pair are first mapped by a shared view-specific transformation:
\begin{equation}
    \tilde{\mathbf{x}}_{i}^{a} = \rho_{a}(\mathbf{x}_{i}^{a}),
    \quad
    \tilde{\mathbf{x}}_{j}^{a} = \rho_{a}(\mathbf{x}_{j}^{a}),
\end{equation}
where $\mathbf{x}_{i}^{u}=\mathbf{u}_{i}$ and $\mathbf{x}_{i}^{v}=\mathbf{v}_{i}$.
The same transformation $\rho_a(\cdot)$ is shared by the two sides of the ordered pair, forming a Siamese scoring structure.
The view-specific pair representation is then constructed as:
\begin{equation}
    \tilde{\mathbf{r}}_{ij}^{a} = [ \tilde{\mathbf{x}}_{i}^{a}; \tilde{\mathbf{x}}_{j}^{a};\tilde{\mathbf{x}}_{i}^{a} - \tilde{\mathbf{x}}_{j}^{a}; \tilde{\mathbf{x}}_{i}^{a}\odot\tilde{\mathbf{x}}_{j}^{a}],
\end{equation}
and the corresponding logit and probability are computed by:
\begin{equation}
    s_{ij}^{a} = \mathbf{w}_{a}^{\top}\tilde{\mathbf{r}}_{ij}^{a}+b_a,
    \quad
    p_{ij}^{a} = \sigma(s_{ij}^{a}),
    \quad a\in\{u,v\}.
\end{equation}
The fused branch uses the pair-conditioned representation $\mathbf{r}_{ij}$ defined in the previous subsection to produce
\begin{equation}
    s_{ij}^{p} = \operatorname{MLP}_{p}(\mathbf{r}_{ij}),
    \quad
    p_{ij}^{p} = \sigma(s_{ij}^{p}).
\end{equation}

The classification loss is applied to all three branches:
\begin{equation}
\begin{aligned}
    \mathcal{L}_{\mathrm{cls}} = -\frac{1}{|\mathcal{T}|} \sum_{(i,j)\in\mathcal{T}} \sum_{a\in\{u,v,p\}} \Big[&y_{ij}\log p_{ij}^{a} \\
    &+ (1-y_{ij})\log(1-p_{ij}^{a}) \Big].
\end{aligned}
\end{equation}

To encourage the two single-view branches to agree with the stronger fused prediction while avoiding unstable mutual updates, ProPRL further adopts a teacher-detached multi-view consistency loss. The fused branch serves as the teacher, and its logit is stopped from gradient back-propagation:
\begin{equation}
\begin{aligned}
    \mathcal{L}_{\mathrm{con}} = \frac{1}{|\mathcal{T}|} \sum_{(i,j)\in\mathcal{T}} \Big( &\left| \sigma(s_{ij}^{u}/t) - \sigma(\operatorname{sg}(s_{ij}^{p})/t) \right| \\
    &+ \left| \sigma(s_{ij}^{v}/t) - \sigma(\operatorname{sg}(s_{ij}^{p})/t) \right| \Big),
\end{aligned}
\end{equation}
where $t$ is the temperature coefficient and $\operatorname{sg}(\cdot)$ denotes the stop-gradient operation.

The final objective combines classification supervision, multi-view consistency, and the irreversibility constraint:
\begin{equation}
    \mathcal{L} = \mathcal{L}_{\mathrm{cls}} + \beta\mathcal{L}_{\mathrm{con}} + \lambda\mathcal{L}_{\mathrm{irr}},
\end{equation}
where $\beta$ and $\lambda$ control the strengths of consistency learning and irreversibility regularization, respectively.

During inference, ProPRL combines the three branch probabilities by a weighted sum:
\begin{equation}
    p_{ij} = w_u p_{ij}^{u} + w_v p_{ij}^{v} + w_p p_{ij}^{p},
    \quad
    w_u+w_v+w_p=1.
\end{equation}
The weights are selected on the validation set and then fixed for test evaluation.

\begin{table*}[t]
\centering
\small
\setlength{\tabcolsep}{4pt}
\renewcommand{\arraystretch}{1.08}
\begin{tabular}{lccccccccc}
\toprule
\multirow{2}{*}{Method}
& \multicolumn{3}{c}{UCD}
& \multicolumn{3}{c}{LectureBank}
& \multicolumn{3}{c}{MOOC} \\
\cmidrule(lr){2-4}
\cmidrule(lr){5-7}
\cmidrule(lr){8-10}
& ACC & F1 & AUC
& ACC & F1 & AUC
& ACC & F1 & AUC \\
\midrule
\rowcolor{gray!8}
\addlinespace[2pt]
\multicolumn{10}{c}{\textit{General-purpose Methods}} \\
\addlinespace[1pt]
\rowcolor{yellow!8}
NB
& 0.5396 & 0.5939 & 0.5039
& 0.5537 & 0.5781 & 0.5156
& 0.5423 & 0.5446 & 0.5511 \\
\rowcolor{yellow!8}
SVM
& 0.4851 & 0.6090 & 0.4841
& 0.4545 & 0.4310 & 0.5068
& 0.5274 & 0.5662 & 0.5625 \\
\rowcolor{yellow!8}
RF
& 0.6436 & 0.6364 & 0.7284
& 0.6777 & 0.6667 & 0.7820
& 0.7214 & 0.7053 & 0.8100 \\
\rowcolor{yellow!8}
RefD
& 0.6386 & 0.6840 & 0.7668
& 0.5041 & 0.6629 & 0.6415
& 0.5821 & 0.6693 & 0.6861 \\
\rowcolor{yellow!8}
GAE
& 0.6634 & 0.7094 & 0.7763
& 0.7438 & 0.7669 & 0.8563
& 0.7910 & 0.7692 & 0.8385 \\
\rowcolor{yellow!8}
VGAE
& 0.6881 & 0.6834 & 0.7714
& 0.7686 & 0.7778 & 0.8525
& 0.7015 & 0.7368 & 0.8312 \\
\rowcolor{gray!8}
\addlinespace[2pt]
\multicolumn{10}{c}{\textit{Prerequisite Learning Methods}} \\
\addlinespace[1pt]
\rowcolor{green!8}
ConLearn
& 0.6275 & 0.7697 & 0.5931
& 0.4771 & 0.4571 & 0.6667
& 0.7961 & 0.8513 & 0.7963 \\
\rowcolor{green!8}
MHAVGAE
& 0.6485 & 0.6844 & 0.7667
& 0.7190 & 0.7344 & 0.7978
& 0.7363 & 0.7135 & 0.8052 \\
\rowcolor{green!8}
HGAPNet
& 0.8069 & 0.7914 & 0.8871
& 0.7934 & 0.7899 & \underline{0.8757}
& 0.8358 & 0.8325 & 0.9024 \\
\rowcolor{green!8}
LCPRE
& 0.7723 & 0.7356 & 0.8655
& 0.7934 & 0.8062 & 0.8730
& 0.8308 & 0.8211 & 0.9000 \\
\rowcolor{green!8}
DGCPL
& \underline{0.8366} & \underline{0.8374} & \underline{0.8934}
& \underline{0.8099} & \underline{0.8099} & 0.8699
& \underline{0.8856} & \underline{0.8832} & \underline{0.9211} \\
\midrule
\rowcolor{red!8}
\textbf{ProPRL (Ours)}
& \textbf{0.8812} & \textbf{0.8788} & \textbf{0.9480}
& \textbf{0.8264} & \textbf{0.8346} & \textbf{0.8929}
& \textbf{0.9254} & \textbf{0.9231} & \textbf{0.9605} \\
\textit{Relative Improv.}
& 5.33\%$\uparrow$ & 4.94\%$\uparrow$ & 6.11\%$\uparrow$
& 2.04\%$\uparrow$ & 3.05\%$\uparrow$ & 1.96\%$\uparrow$
& 4.49\%$\uparrow$ & 4.52\%$\uparrow$ & 4.28\%$\uparrow$ \\
\bottomrule
\end{tabular}
\caption{Overall performance comparison on three benchmark datasets. The best results are highlighted in bold.}
\label{tab:performance_comparison}
\end{table*}

\begin{table}[t]
\centering
\small
\setlength{\tabcolsep}{6pt}
\renewcommand{\arraystretch}{1.08}
\begin{tabular}{lccc}
\toprule
\rowcolor{gray!8}
\textbf{Variant} & \textbf{UCD} & \textbf{LectureBank} & \textbf{MOOC} \\
\midrule
w/o Pair Gate
& 0.8528 & 0.8254 & 0.8763 \\
w/o Multi-hop Prop.
& 0.7831 & 0.8000 & 0.8808 \\
w/o Irreversibility
& 0.8670 & 0.7969 & 0.9109 \\
\midrule
\rowcolor{red!8}
\textbf{ProPRL}
& \textbf{0.8788} & \textbf{0.8346} & \textbf{0.9231} \\
\bottomrule
\end{tabular}
\caption{Ablation study of ProPRL in terms of F1-score.}
\label{tab:ablation}
\end{table}

\section{Experiments}
\paragraph{\textbf{Datasets.}} We evaluate ProPRL on three benchmark datasets for prerequisite relation learning: MOOC\footnote{https://github.com/suderoy/PREREQ-IAAI-19} ~\cite{LiaYW2017}, LectureBank\footnote{https://github.com/Yale-LILY/LectureBank}~\cite{LiFT2019}, and University Course\footnote{https://github.com/suderoy/PREREQ-IAAI-19}(UCD)~\cite{LiaYW2017}.
%These datasets cover different educational scenarios and contain concept sets, learning resources, learner behavior transitions, and labeled prerequisite concept pairs.

\paragraph{\textbf{Evaluation Metrics.}} We use accuracy (ACC), F1-score (F1), and area under the ROC curve (AUC) as the main evaluation metrics, following previous studies on prerequisite relation learning~\cite{ZhaWH2025}. For all metrics, higher values indicate better performance.

\paragraph{\textbf{Baseline Methods.}} 
We compare ProPRL with two groups of baselines: general-purpose methods and task-specific prerequisite relation learning models. The first group includes NB, SVM, RF, RefD~\cite{LiaWH2015}, GAE~\cite{KipW2016}, and VGAE~\cite{KipW2016}. The second group includes HGAPNet~\cite{MazPB2023}, MHAVGAE~\cite{ZhaLZ2022}, ConLearn~\cite{SunLZ2022}, LCPRE~\cite{SunHX2024}, and DGCPL~\cite{ZhaWH2025}.

\paragraph{\textbf{Implementation Details.}}
ProPRL is implemented with PyTorch.
The datasets are split into training, validation, and test sets at a ratio of 8:1:1.
For all datasets, the number of training epochs, batch size, learning rate, $\mu$, $t$, $\beta$ and seed are fixed at 50, 16, $1\times10^{-4}$, 0.8, 0.5, $1\times10^{-5}$ and 42, respectively.
We set $\alpha$ to $0.05$ for LectureBank and MOOC and to $0.2$ for UCD.
Moreover, $\lambda$ is set to $1\times10^{-3}$ for UCD and MOOC and to $5\times10^{-3}$ for LectureBank.
All experiments are conducted using an NVIDIA GeForce RTX 3070 Ti GPU and an AMD Ryzen 5800x CPU.

\paragraph{\textbf{Performance Comparison.}}
Table~\ref{tab:performance_comparison} presents the overall performance comparison on the three benchmark datasets. The most notable observation is the consistency of ProPRL, which ranks first across all nine dataset–metric combinations, although the identity of the strongest baseline varies across datasets and metrics.

Specifically, DGCPL constitutes the strongest baseline in eight of the nine comparisons, whereas HGAPNet achieves the highest baseline AUC on LectureBank. ProPRL nevertheless outperforms the strongest baseline in every comparison, demonstrating a consistent performance advantage across different evaluation settings.
The improvements are observed jointly in ACC and F1, which characterize overall classification accuracy and the balance between precision and recall, as well as in AUC, which evaluates ranking quality independently of a specific decision threshold. Notably, ProPRL improves AUC over the strongest baseline on all three datasets, with the largest relative gain of 6.11\% observed on UCD. Overall, the relative improvements range from 1.96\% to 6.11\% across the nine comparisons, demonstrating the effectiveness and cross-dataset consistency of ProPRL.

\paragraph{\textbf{Ablation Study.}}
Table~\ref{tab:ablation} evaluates the contribution of the three principal components of ProPRL. Removing any component consistently reduces the F1-score across all three datasets, confirming that each component contributes positively to the complete model. Among them, multi-hop propagation has the largest overall impact: its removal causes the most pronounced degradation on UCD, decreasing the F1-score from 0.8788 to 0.7831. This result highlights the importance of propagating behavioral evidence beyond immediate neighbors for prerequisite relation learning.

The effects of the other components vary across datasets. Removing the pair-aware gate produces the largest degradation on MOOC, while removing anti-symmetry regularization affects LectureBank most strongly. These dataset-dependent patterns indicate that relation-specific evidence fusion and directional constraints address distinct aspects of prerequisite relation learning. Taken together, the consistent degradation of all ablated variants validates the necessity of the three components in the complete ProPRL framework.

\begin{figure}[t]
  \centering
  \includegraphics[width=\linewidth]{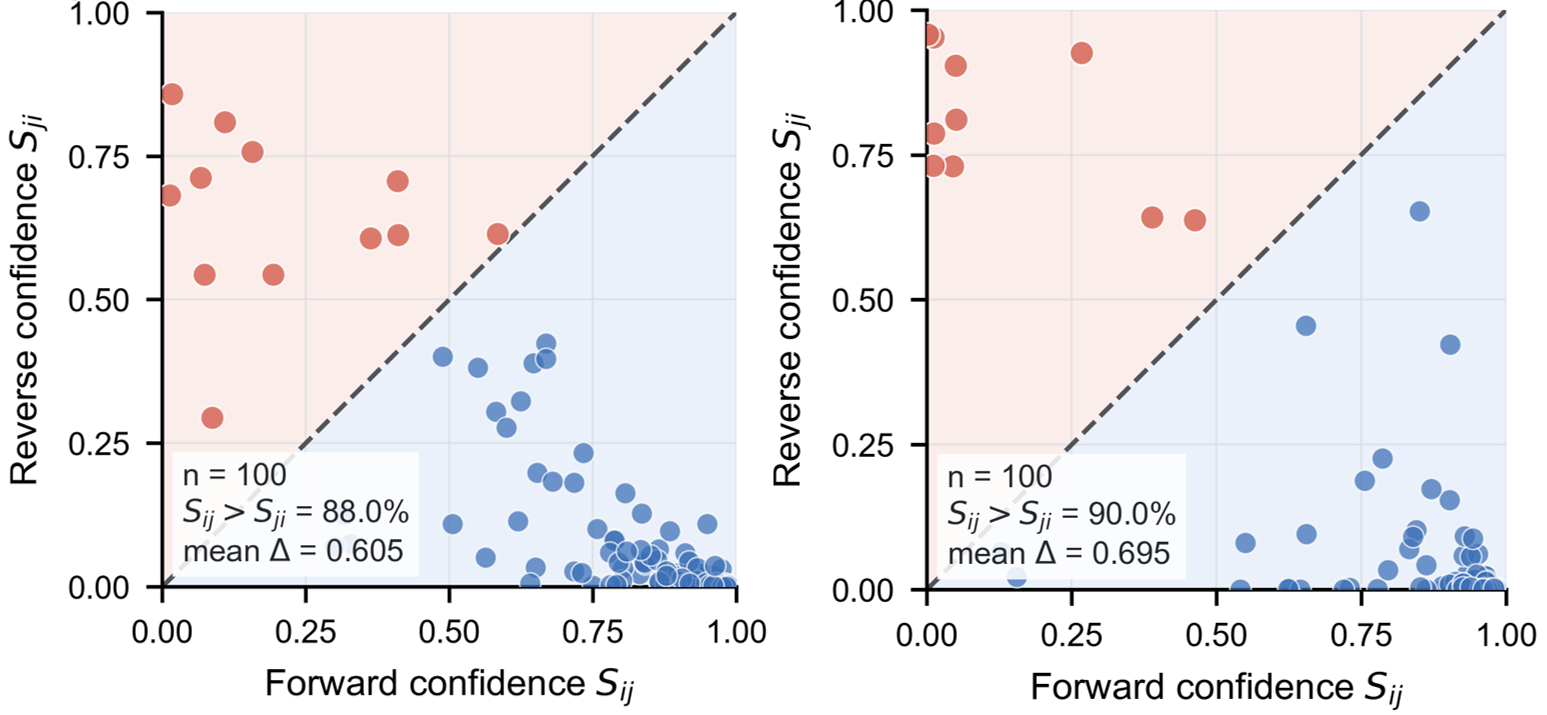}
  \caption{The left panel shows the results of DGCPL, while the right panel shows the results of ProPRL. Each point denotes a positive test relation, with the annotated-direction confidence $S_{ij}$ and reversed-direction confidence $S_{ji}$. }
  \label{fig:case}
\end{figure}

\paragraph{\textbf{Case Study.}}
To examine whether the models distinguish a prerequisite relation from its reversed counterpart, we evaluated each positive MOOC test relation in both orders. As shown in Figure~\ref{fig:case}, both models place most relations below the diagonal, indicating that they generally assign higher confidence to the annotated prerequisite direction. ProPRL, however, produces a clearer concentration toward the lower-right region, where the annotated direction receives high confidence while the reversed direction is strongly suppressed.

Quantitatively, ProPRL increases the proportion of correctly ordered relations from 88.0\% to 90.0\% and enlarges the mean forward-reverse confidence margin from 0.605 to 0.695. The wider margin results from both a higher mean confidence for the annotated direction and a lower mean confidence for its reversal. Therefore, the main improvement is not merely the correction of a small number
of reversed rankings, but a stronger separation between the two possible directions across the test relations.

\begin{figure}[t]
  \centering
  \includegraphics[width=\linewidth]{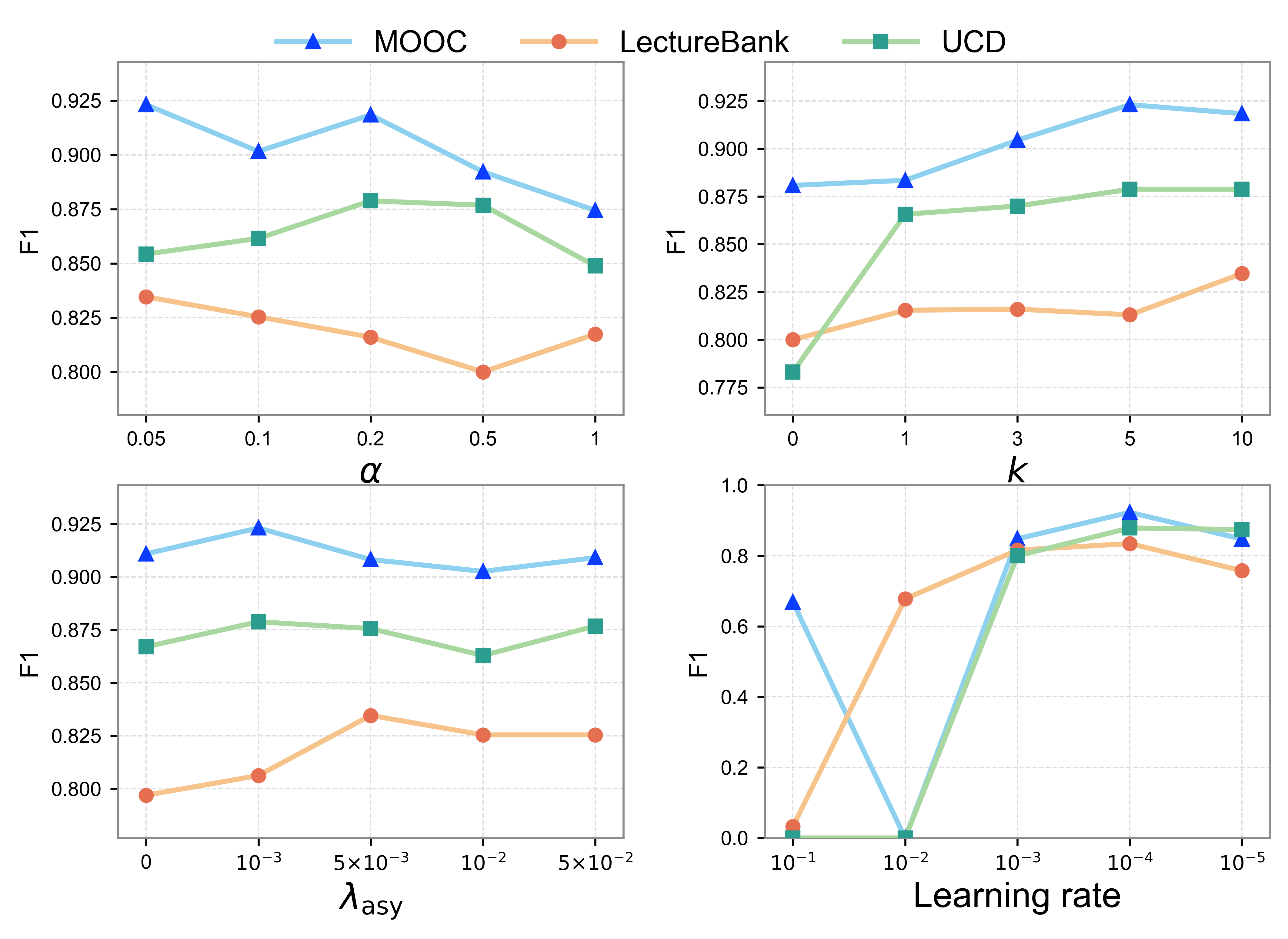}
  \caption{Hyperparameter Study.}
  \label{fig:hyper}
\end{figure}

\paragraph{\textbf{Hyperparameter Study.}}
Figure~\ref{fig:hyper} investigates the sensitivity of ProPRL to the propagation coefficient $\alpha$, propagation depth $k$, anti-symmetry weight $\lambda_{\mathrm{asy}}$, and learning rate. The effect of $\alpha$ is dataset-dependent, although ProPRL maintains relatively competitive performance across the tested range. MOOC favors a smaller $\alpha$, whereas UCD performs best at a moderate value, indicating that the appropriate balance between propagated information and the initial representation depends on the underlying graph structure.

Increasing $k$ from zero generally improves performance, with particularly pronounced gains on UCD. The improvements become limited once $k$ reaches approximately 5, suggesting that a moderate propagation depth captures most of the useful multi-hop behavioral dependencies, while further propagation provides little additional benefit. ProPRL is comparatively insensitive to $\lambda_{\mathrm{asy}}$ over the evaluated range: nonzero regularization usually improves or maintains performance, but the optimal strength varies across datasets. This shows that directional regularization is beneficial without requiring a narrowly tuned coefficient.

In contrast, the learning rate has a substantially stronger effect on model performance. The best overall performance is obtained around $10^{-4}$. Taken together, these results indicate that ProPRL is reasonably stable with respect to its structural hyperparameters but requires an appropriately small learning rate for reliable optimization.

\begin{table}[t]
\centering
\footnotesize
\setlength{\tabcolsep}{4.2pt}
\renewcommand{\arraystretch}{1.08}
\begin{tabular}{@{}llrrrr@{}}
\toprule
\rowcolor{gray!8}
Data & Method & Params & Train & Infer. & Mem. \\
\rowcolor{gray!8}
     &        & (M)    & (s)   & (s)    & (MB) \\
\midrule
\multirow{2}{*}{MOOC}
& DGCPL & 1.89 & 122.44 & 0.0707 & 28.50 \\
& \cellcolor{red!8} ProPRL  & \cellcolor{red!8} 2.68 & \cellcolor{red!8} 211.59 & \cellcolor{red!8} 0.0316 & \cellcolor{red!8} 37.35 \\
\addlinespace[1pt]
\midrule
\multirow{2}{*}{LectureBank}
& DGCPL & 1.89 & 73.69  & 0.0427 & 26.48 \\
& \cellcolor{red!8} ProPRL  & \cellcolor{red!8} 2.68 & \cellcolor{red!8} 96.95  & \cellcolor{red!8} 0.1754 & \cellcolor{red!8} 34.11 \\
\addlinespace[1pt]
\midrule
\multirow{2}{*}{UCD}
& DGCPL & 1.89 & 127.39 & 0.0735 & 28.73 \\
& \cellcolor{red!8} ProPRL  & \cellcolor{red!8} 2.68 & \cellcolor{red!8} 212.01 & \cellcolor{red!8} 0.0351 &  \cellcolor{red!8} 37.82 \\
\bottomrule
\end{tabular}
\caption{Efficiency comparison.}
\label{tab:efficiency}
\end{table}

\paragraph{\textbf{Efficiency Study.}}
Table~\ref{tab:efficiency} reports the computational efficiency of ProPRL and DGCPL on the original data splits. Compared with DGCPL, ProPRL incurs a moderate increase in model size and training time because it learns complementary representations from the concept-resource and learning-behavior views and performs pair-conditioned gated fusion. The direction-preserving personalized propagation and the irreversibility constraint further introduce additional training overhead. Nevertheless, ProPRL remains lightweight, using less than 40 MB of GPU memory on all datasets and requiring less than 0.18 seconds for test-time inference. ProPRL is faster than DGCPL on MOOC and UCD, although it requires more inference time on LectureBank. Overall, these results show that ProPRL achieves improved prerequisite relation modeling with acceptable computational overhead.

\section{Conclusion}
In this paper, we presented \textbf{ProPRL}, a property-aware framework for prerequisite relation learning in educational knowledge graphs. ProPRL learns complementary concept representations from a concept-resource hypergraph and a directed learning-behavior graph, employs a Pair-conditioned Gate to adaptively fuse evidence for each candidate ordered concept pair, and incorporates an anti-symmetry regularizer to discourage simultaneously high confidence in reverse directions. Experiments on three benchmark datasets showed that ProPRL consistently outperformed representative baselines, while ablation studies further supported the effectiveness of its principal components.

\bibliography{aaai2027}


\end{document}